\documentclass{article}

\usepackage{arxiv}

\usepackage[utf8]{inputenc} % allow utf-8 input
\usepackage[T1]{fontenc}    % use 8-bit T1 fonts
\usepackage{hyperref}       % hyperlinks
\usepackage{url}            % simple URL typesetting
\usepackage{booktabs}       % professional-quality tables
\usepackage{amsfonts}       % blackboard math symbols
\usepackage{nicefrac}       % compact symbols for 1/2, etc.
\usepackage{microtype}      % microtypography
\usepackage{lipsum}
\usepackage{graphicx}
\graphicspath{ {./images/} }
\usepackage{booktabs}

\title{Spanish Legalese Language Model and Corpora}

\author{
 Asier Gutiérrez-Fandiño \\
  Text Mining Unit\\
  Barcelona Supercomputing Center\\
  \texttt{asier.gutierrez@bsc.es} \\
\And
   Jordi Armengol-Estapé\\
  Text Mining Unit\\
  Barcelona Supercomputing Center\\
  \texttt{jordi.armengol@bsc.es} \\
\And
   Aitor Gonzalez-Agirre\\
  Text Mining Unit\\
  Barcelona Supercomputing Center\\
  \texttt{aitor.gonzalez@bsc.es} \\
\And
   Marta Villegas\\
  Text Mining Unit\\
  Barcelona Supercomputing Center\\
  \texttt{marta.villegas@bsc.es} \\
}

\begin{document}
\maketitle
\begin{abstract}
There are many Language Models for the English language according to its worldwide relevance. However, for the Spanish language, even if it is a widely spoken language, there are very few Spanish Language Models which result to be small and too general. Legal slang could be think of a Spanish variant on its own as it is very complicated in vocabulary, semantics and phrase understanding. For this work we gathered legal-domain corpora from different sources, generated a model and evaluated against Spanish general domain tasks. The model provides reasonable results in those tasks. % both general domain tasks and domain-specific tasks. The model provides reasonable results in general domain tasks and it excels in domain-specific tasks.

\end{abstract}

\section{Introduction}
Legal Spanish (or Spanish Legalese) is a complex slang that is away from the language spoken by the society.

Language Models, generally, are pre-trained on large corpora for later fine-tuning them on different tasks. Language Models are widely used due to their transfer learning capabilities. If the corpora used for training the Language Models are aligned with the domain of the tasks they provide better results.

In this work we gathered different corpora and we trained a Language Model for the Spanish Legal domain.

\section{Corpora}
Our corpora comprises multiple digital resources and it has a total of 8.9GB of textual data. Part of it has been obtained from previous work \cite{samy-etal-2020-legal}. Table \ref{tab:corpora} shows different resources gathered. Most of the corpora were scraped, some of them in PDF format. We then transformed and cleaned the data. Other corpora like the COPPA\footnote{\url{https://www.wipo.int/export/sites/www/patentscope/en/data/pdf/wipo-coppa-technicalDocumentation.pdf}} patents corpus were requested.

As a contribution of this work we publish all publishable corpora we gathered in Zenodo\footnote{\url{https://zenodo.org/record/5495529}}.
\begin{table}[h]
\centering
\begin{tabular}{@{}lrr@{}}
\toprule
Corpus name & \multicolumn{1}{l}{Size (GB)} & \multicolumn{1}{l}{Tokens (M)} \\ \midrule
Procesos Penales & 0.625 & 0.119 \\
JRC Acquis & 0.345 & 59.359 \\
Códigos Electrónicos Universitarios & 0.077 & 11.835 \\
Códigos Electrónicos & 0.080 & 12.237 \\
Doctrina de la Fiscalía General del Estado & 0.017 & 2.669 \\
Legislación BOE & 3.600 & 578.685 \\
Abogacía del Estado BOE & 0.037 & 6.123 \\
Consejo de Estado: Dictámenes & 0.827 & 135.348 \\
Spanish EURLEX & 0.001 & 0.072 \\
UN Resolutions & 0.023 & 3539.000 \\
Spanish DOGC & 0.826 & 132.569 \\
Spanish MultiUN & 2.200 & 352.653 \\
Consultas Tributarias Generales y Vinculantes & 0.466 & 77.691 \\
Constitución Española & 0.002 & 0.018 \\
COPPA Patents Corpus & 0.002 & - \\
Biomedical Patents & 0.083 & - \\ \bottomrule
\end{tabular}
\caption{List of individual sources constituting the legal corpus. The number of tokens refers to \emph{white-spaced} tokens calculated on cleaned untokenized text.}
\label{tab:corpora}
\end{table}

\section{Model}
We trained a RoBERTa \cite{liu2019roberta} base model, using the hyper-parameters proposed in the original work. As vocabulary, we used Byte-Level BPE or training, we use the Fairseq \cite{ott2019fairseq} library, and for fine-tuning, Huggingface Transformers \cite{wolf-etal-2020-transformers}, but with a vocabulary size of 52,262. For training, we used the Fairseq \cite{ott2019fairseq} library, and for fine-tuning, Huggingface Transformers \cite{wolf-etal-2020-transformers}. We trained the model until convergence with 8 Nvidia Tesla V100 GPUs with 16GB of VRAM. The model was trained with a peak learning rate of 0.0005 and 2,048 of batch size. The model is available in HuggingFace.\footnote{\url{https://huggingface.co/BSC-TeMU/RoBERTalex}}

\section{Embeddings}
Additionally, following previous work in the Natural Language Processing for Spanish Legal texts \cite{samy-etal-2020-legal} we computed FastText word and subword embeddings, with 50, 100 and 300 dimensions, using CBOW and Skip-gram methods. For the word embeddings, we computed both cased and uncased word embeddings, and for the subword embeddings we computed Byte-level Byte-Pair-Encoding (BBPE) embeddings with 30k vocabulary size. The embeddings can be freely downloaded from Zenodo\footnote{\url{https://zenodo.org/record/5036147}}.

\section{Evaluation}

We compare our RoBERTalex model with the Spanish RoBERTa-base (RoBERTa-b) \cite{gutierrezfandino2021spanish} and multilingual BERT (mBERT) \cite{DBLP:journals/corr/abs-1810-04805}. Due to the lack of domain specific evaluation data, he models are evaluated on general domains tasks, where RoBERTalex obtains reasonable performance. We fine-tuned each model in the following tasks:

\begin{itemize}
    \item Part of Speech from Universal Dependencies\footnote{\url{https://universaldependencies.org/}} (UD-POS).
    \item Named Entity Recognition from Conll2002 (Conll-NER) \cite{tjong-kim-sang-2002-introduction}.
    \item Part of Speech from the Capitel Corpus (Capitel-POS).\footnote{\url{https://sites.google.com/view/capitel2020\#h.p_eFTF8UCJXFMq}}
    \item Named Entity Recognition from the Capitel Corpus (Capitel-NER).\footnote{\url{https://sites.google.com/view/capitel2020\#h.p_CbqX2kG3XEIp}}
    \item Semantic Textual Similarity (STS) from 2014 \cite{agirre2014semeval} and 2015 \cite{agirre2015semeval}.
    \item The Multilingual Document Classification Corpus (MLDoc) \cite{mldoc, reuters}.
    \item The Cross-lingual Adversarial Dataset for Paraphrase Identification (PAWS-X) \cite{pawsx}.
    \item The Cross-Lingual NLI Corpus (XNLI) \cite{xnli}.
    % aquí tendríamos que poner el SQAC!
\end{itemize}

Table \ref{tab:evaluation_general} shows the evaluation results of the three models. RoBERTalex was evaluated with a fixed set of hyper-parameters, while the results reported in \cite{gutierrezfandino2021spanish} were obtained by conducting a grid search and picking the best value based on the development set. We plan to evaluate RoBERTalex using the same grid search in order to make the results fully comparable, and also we plan to evaluate the model in domain-specific tasks.

\begin{table}[h!]
\centering
\begin{tabular}{@{}llrrrrr@{}}
\toprule
Dataset & Metric & \multicolumn{1}{l}{RoBERTalex} & \multicolumn{1}{l}{RoBERTa-b} & \multicolumn{1}{l}{mBERT} \\ \midrule
UD-POS & F1 & 0.9871 & {\bf 0.9907} & 0.9886 \\
Conll-NER & F1 & 0.8323 & {\bf 0.8851} & 0.8691 \\
Capitel-POS & F1 & 0.9788 & {\bf 0.9846} & 0.9839 \\
Capitel-NER & F1 & 0.8394 & {\bf 0.8960} & 0.8810 \\
STS & Combined & 0.7374 & {\bf 0.8533} & 0.8164 \\
MLDoc & Accuracy & 0.9417 & {\bf 0.9623} & 0.9550 \\
PAWS-X & F1 & 0.7304 & {\bf 0.9000} & 0.8955 \\
XNLI & Accuracy & 0.7337 & {\bf 0.8016} & 0.7876 \\ \bottomrule
\end{tabular}
\caption{Evaluation table of models.}
\label{tab:evaluation_general}
\end{table}

\section{Conclusions \& Future Work}
Our language model is, to our knowledge, the first of its kind (Spanish legal domain). We extensively evaluated our model by performing general domain evaluation. Results show that it behaves reasonably positive in general domain.

We are planning to gather more resources, try to continue the pre-training of the models from \cite{gutierrezfandino2021spanish} with legal domain and train generative models.

\section*{Acknowledgements}
This work was funded by the Spanish State Secretariat for Digitalization and Artificial Intelligence (SEDIA) within the framework of the Plan-TL.

\bibliographystyle{abbrv}  

\bibliography{references}

\end{document}